\pdfoutput=1

\documentclass[11pt]{article}

\usepackage[]{acl}

\usepackage{times}
\usepackage{latexsym}

\usepackage[T1]{fontenc}

\usepackage[utf8]{inputenc}

\usepackage{microtype}

\usepackage{multirow}
\usepackage{amsmath}
\usepackage{caption}
\usepackage{array}
\usepackage{booktabs}
\usepackage{footnote}
\usepackage{graphicx}
\usepackage{color}
\usepackage{subfig}
\usepackage{amssymb}
\usepackage[ruled, linesnumbered]{algorithm2e}
\usepackage{threeparttable}
\usepackage{booktabs}
\usepackage[switch]{lineno}

\usepackage{enumerate}
\usepackage{enumitem}
\usepackage{ulem}
\usepackage{verbatim}
\usepackage{colortbl}

\newcommand{\name}{{TMEG}}

\usepackage{ulem}

\title{Modeling Temporal-Modal Entity Graph for Procedural \par Multimodal Machine Comprehension}

\author{
  Huibin Zhang$^1$, Zhengkun Zhang$^1$, Yao Zhang$^1$, Jun Wang$^2$\footnotemark[1], Yufan Li$^1$ \\ {\bf Ning Jiang$^3$, Xin Wei$^3$, Zhenglu Yang$^{1}$\footnotemark[1]}\\
  $^1$TKLNDST, CS, Nankai University, China, $^2$Ludong University, China, \\
  $^3$Mashang Consumer Finance Co., Ltd., China \\
  \texttt{\{zhanghuibin.cc,zhangzk2017,yaozhang,junwang,1811486\}} \\ 
  \texttt{@mail.nankai.edu.cn,\{ning.jiang02,xin.wei02\}@msxf.com,} \\
  \texttt{yangzl@nankai.edu.cn}
  }

\begin{document}
\maketitle
\renewcommand{\thefootnote}{\fnsymbol{footnote}} %
\footnotetext[1]{Corresponding authors.} %
\renewcommand{\thefootnote}{\arabic{footnote}} %

\begin{abstract}

Procedural Multimodal Documents (PMDs) %
organize textual instructions and corresponding images step by step.
Comprehending PMDs and inducing their representations for the downstream reasoning tasks %
is designated as Procedural MultiModal Machine Comprehension (M$^3$C). %
In this study, we approach Procedural M$^3$C at a fine-grained level (compared with existing explorations at a document or sentence level), that is, entity. 
With delicate consideration, we model entity both in its temporal and cross-modal relation and propose
a novel Temporal-Modal Entity Graph (\name{}). %
Specifically, a heterogeneous graph structure is formulated to capture textual and visual entities and trace their temporal-modal evolution.
In addition, a graph aggregation module is introduced to conduct graph encoding and reasoning.
Comprehensive experiments across three Procedural M$^3$C tasks are conducted on a traditional dataset RecipeQA and our new dataset CraftQA, which %
can better evaluate the generalization of \name{}. 

\end{abstract}

\section{Introduction}

\textbf{M}ulti\textbf{M}odal \textbf{M}achine \textbf{C}omprehension (M$^3$C) is a generalization of machine reading comprehension %
by introducing multimodality.
Due to its differences from Visual Question Answering (VQA)~\cite{DBLP:conf/iccv/AntolALMBZP15} in the form of understanding multimodal contexts and conducting multimodal questions and answers, there has been a lot of attention in recent years devoted to this field.
In this paper, we investigate a task that has been typical of M$^3$C recently, named Procedural M$^3$C, a task of reading comprehension of \textbf{P}rocedural \textbf{M}ultimodal \textbf{D}ocuments~(PMDs).

\begin{figure}
\centering
\includegraphics[width=.49\textwidth]{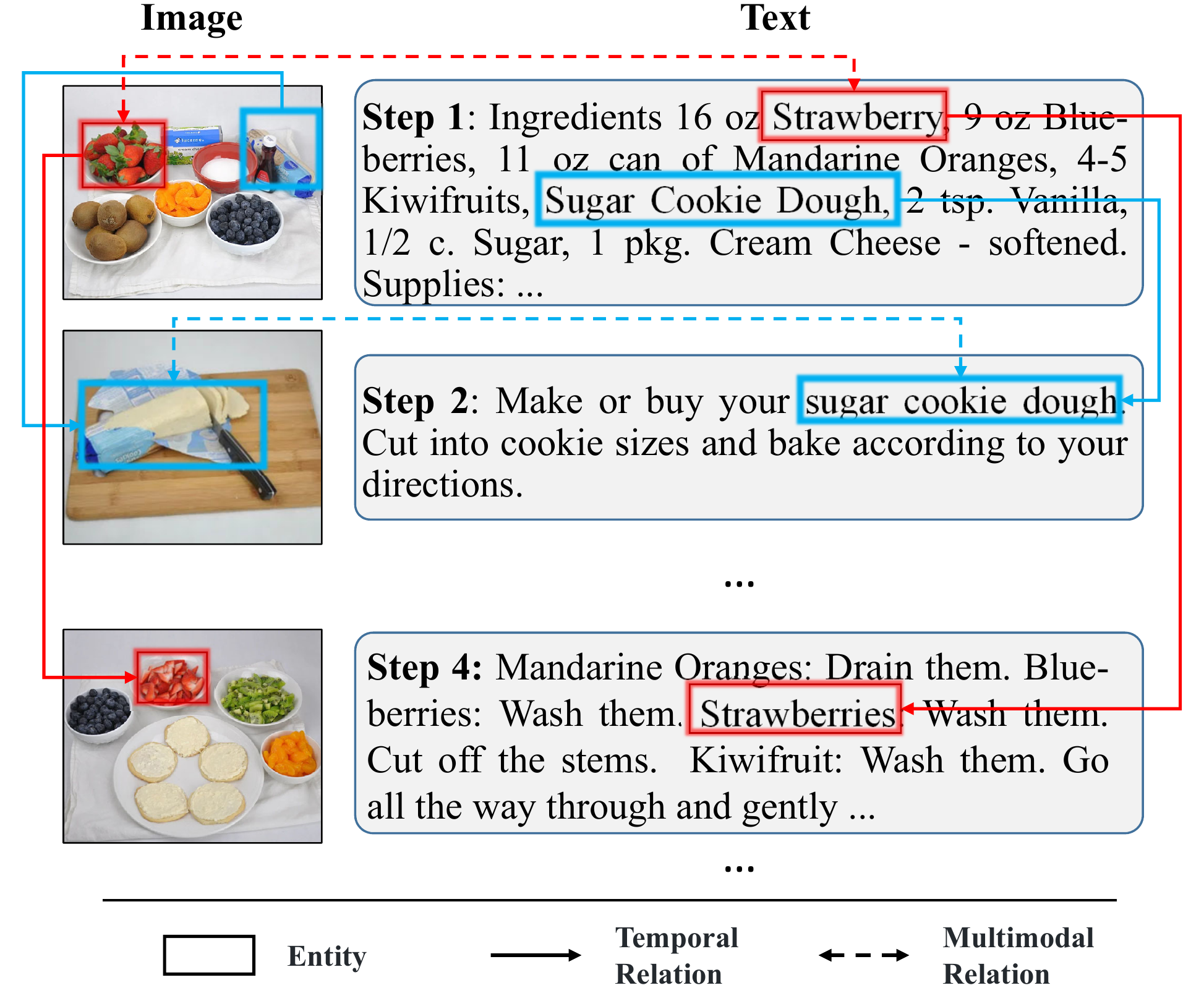}
\caption{
A PMD instance: a recipe about making mini fruit pizza.
The entities and relations (i.e. temporal and multimodal) among them are highlighted.
}
\label{fig:recipe}
\end{figure}

As shown in Figure~\ref{fig:recipe}, a recipe that contains successive multimodal instructions is a typical PMD. 
Reading a recipe seems trivial for humans but is still %
complex for a machine reading comprehension system before it can comprehend both textual and visual contents and capture their relations.

Current Procedural M$^3$C studies~\cite{DBLP:conf/emnlp/YagciogluEEI18,DBLP:conf/sigir/LiuYZLLBX20} comprehend PMDs by encoding text and images at each procedure step.
These efforts, however, only scratch the surface and %
lack deep insight into the elementary unit of PMDs, that is, 
\textbf{entity}.
From now on, we use entity to refer uniformly to entity in text and object in image.
For instance, the recipe in Figure~\ref{fig:recipe} involves multiple entities, i.e., \textsf{Strawberry} and \textsf{Sugar Cookie Dough}, etc. 
In this work, we target at approaching the Procedural M$^3$C task at a fine-grained entity level.

We observe that a PMD essentially assembles an evolution process of entities
and the relations between them.
Specifically, the relation between entities
can be summarized in the following two categories:

\begin{itemize}
[leftmargin=*]
\setlength{\itemsep}{0pt}
\setlength{\parsep}{0pt}

\item[\tiny$\bullet$] \textbf{Temporal Relation.} The state of an entity %
may change as steps progress. Still looking at Figure~\ref{fig:recipe}, \textsf{strawberries} are complete at step 1 and by step 4 they are washed and cut into pieces. 
We use temporal relation to depict the association between an entity's changing visual signals in images or changing contexts in text.

\item[\tiny$\bullet$] \textbf{Multimodal Relation.} An entity is naturally and powerfully associated with other entities within a single modality. 
Meanwhile, the cross-modal association of an entity is worth exploiting to contribute to distinct modality understanding.
For example, the visual signal and context about \textsf{sugar cookie dough} can be interpreted by each other at step 2.
We generalize the intra- and inter- modal associations of entities with the multimodal relation.
\end{itemize}

Based on the above observations, we believe that simultaneously modeling entities and the temporal and multimodal relations is a key challenge in understanding PMDs.
Recent efforts~\cite{DBLP:conf/conll/AmacYEE19,DBLP:conf/acl/HuangGPLJ20} are devoted to encoding temporal relations of entities, while it neglects the multimodal relation.
Heterogeneous graphs have become the preferred technology for representing, sharing, and fusing information to modern AI tasks, e.g., relation extraction~\cite{christopoulou2019connecting} and recommendation~\cite{fan2019metapath}.
Inspired by this, we construct a heterogeneous graph, with entities as nodes and relations as edges.
Therefore, the research goal of the procedural M$^3$C task will shift from understanding unstructured PMDs to learning structured graph representations.

In this work, we propose a novel \textbf{T}emporal \textbf{M}odal \textbf{E}ntity \textbf{G}raph model, namely \textbf{\name{}}. 
Our model approaches Procedural M$^3$C at a fine-grained entity level by constructing and learning a graph with entities and temporal and multimodal relations.
Specifically, \name{} consists of the following components:
1) \textbf{Node Construction}, which extracts the token-level features in text and object-level features in images as the initial entity embeddings;
2) \textbf{Graph Construction}, which constructs the temporal, intra-modal, and cross-modal relations separately to form a unified graph;
and 3) \textbf{Graph Aggregation}, which utilizes the graph-based multi-head attention mechanism to perform fusion operations on graphs to model the evolution of entities and relations.
Finally, the graph representation is fed into a graph-based reasoning module to evaluate the model's understanding ability.

In addition, in order to further advance the research of Procedural M$^3$C, we release CraftQA, a multimodal semantically enriched dataset that contains about 27k craft product-making tutorials and 46k question-answer pairs for evaluation.
We evaluate three representative subtasks, i.e., visual cloze, visual coherence, and visual ordering, on CraftQA and a public dataset RecipeQA~\cite{DBLP:conf/emnlp/YagciogluEEI18}.
The quantitative and qualitative results show the superiority of \name{} compared to the state-of-the-art methods on all three tasks.
The main contributions of this paper can be summarized as follows:

\begin{itemize}
[leftmargin=*]
\setlength{\itemsep}{0pt}
\setlength{\parsep}{0pt}

\item[\tiny$\bullet$] We innovatively study the Procedural M$^3$C task at a fine-grained entity level. We comprehensively explore the relations between entities in both temporal and multimodal perspectives.
\item[\tiny$\bullet$] We propose a Temporal-Modal Entity Graph model \name{}, which constructs a graph with entities as nodes and relations as edges, and then learns the graph representation to understand PMDs.
\item[\tiny$\bullet$] We release a dataset CraftQA. The experimental results on CraftQA and RecipeQA show \name{} outperforms several state-of-the-art methods.

\end{itemize}

\begin{figure*}
  \centering
  \includegraphics[width=1\textwidth]{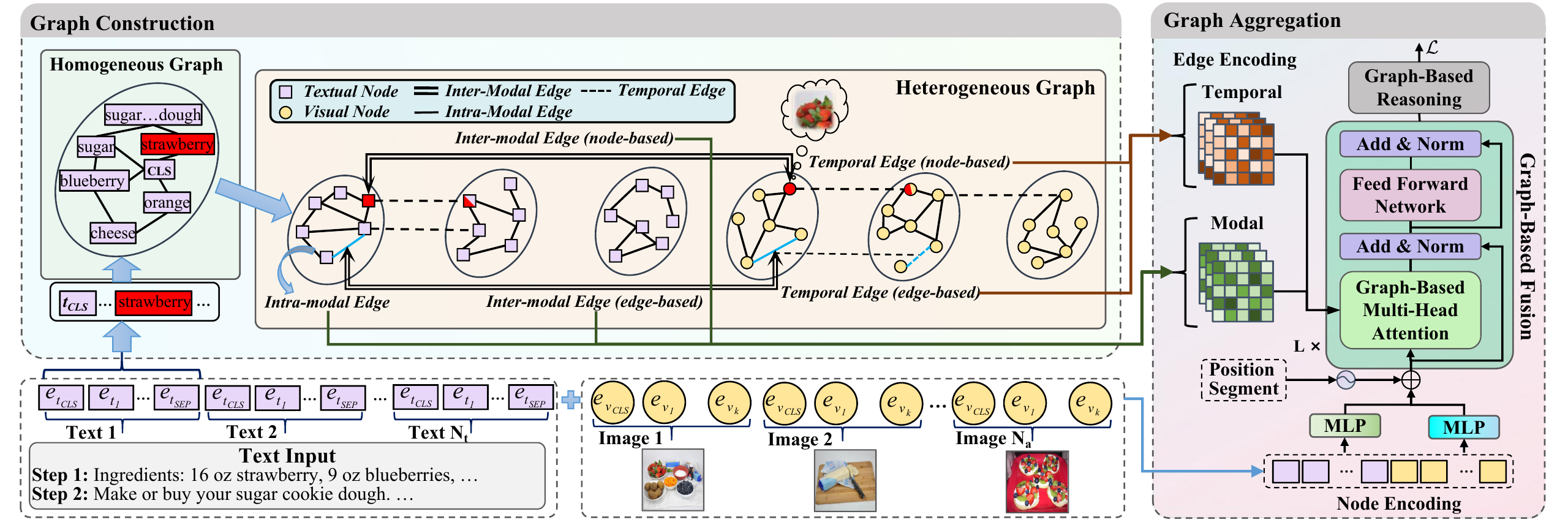} \\
  \caption{
  Overview of our proposed \name{} framework. 
We reuse the example in Figure~\ref{fig:recipe}. 
Initial nodes are generated from input text and images. 
Considering the temporal and cross-modal relation of entities, we apply various types of edges to construct a unified temporal-modal entity graph.
By combining the encoding of nodes and edges, a graph aggregation module is designed to conduct graph encoding and reasoning.
 }
  \label{fig:framework}
\end{figure*}

\section{Related Work}

\noindent \textbf{Procedural Text Comprehension.}
Procedural text comprehension requires %
an accurate prediction for the state change and location information of each entity over successive steps.
Several datasets have been proposed to evaluate procedural text comprehension, e.g., Recipe~\cite{DBLP:conf/iclr/BosselutLHEFC18}, ProPara~\cite{DBLP:journals/corr/abs-1909-04745}, and OPENPI~\cite{DBLP:conf/emnlp/TandonSDRCGRH20}.
To model %
entity evolution,
KG-MRC~\cite{DBLP:conf/iclr/DasMYTM19} constructs knowledge graphs via reading comprehension and uses them for entity location prediction. 
DYNAPRO~\cite{DBLP:conf/akbc/AminiBMCH20} introduces a pre-trained language model to dynamically obtain the contextual embedding of procedural text and learn the attributes and transformations of entities. 
ProGraph~\cite{DBLP:journals/corr/abs-2004-12057} enables state prediction from context by constructing a heterogeneous graph with various knowledge inputs. 
KoalA~\cite{DBLP:conf/www/ZhangGQWJ21} utilizes the external commonsense knowledge injection and data enhancement to reason the states and locations of entities.
TSLM~\cite{DBLP:conf/naacl/FaghihiK21} formulate comprehension task as a question answering problem and adapt pre-trained transformer-based language models on other QA benchmarks.
REAL~\cite{DBLP:conf/acl/HuangGPLJ20} builds a general framework to systematically model the entity, action, location by using a graph neural network.

Inspired by these previous works, we propose a temporal-modal entity graph model, which is designed with temporal encoding and modal encoding to model multiple types of entities.

\textbf{Multimodal Graph.}
In recent multimodal research, graph structure has been utilized to model the semantic interaction between modalities.
\cite{DBLP:conf/acl/YinMSZYZL20} propose a graph-based multimodal fusion encoder for neural machine translation, which converted sentence and image in a unified multimodal graph.
\cite{DBLP:conf/acl/Khademi20} convert image regions and the region grounded captions into graph structure and introduced graph memory networks for visual question answering.
\cite{DBLP:conf/aaai/ZhangWLWZZ21} propose a multimodal graph fusion approach for named entity recognition, which conducted graph encoding via multimodal semantic interaction.
\cite{DBLP:conf/naacl/YangWYZRZPM21} focus on multimodal sentiment analysis and emotion recognition, which %
unified video, audio, and text modalities into an attention graph and learned the interaction %
through graph fusion, dynamic pruning, and the read-out technique.

In contrast to the above methods, %
we formulate our multimodal graph on temporal entities and successfully deploy it in Procedural M$^3$C.

\section{Proposed Method}\label{sec:proposed method}
In this section, we %
introduce: (1) problem definition of Procedural M$^3$C in Section~\ref{sec:problem definition}; 
(2) The homogeneous graph of each textual instruction (image) and our \name{} in Section~\ref{sec:graph construction} and Section~\ref{sec:edge construction}, respectively;
and (3) graph aggregation module to conduct graph encoding and reasoning in Section~\ref{sec:graph aggregation}.
Figure~\ref{fig:framework} gives a high-level overview of %
\name{}.

\subsection{Problem Definition}\label{sec:problem definition}
\label{Problem Definition}
Here, we define the task of Procedural M$^3$C, given:
\begin{itemize}[itemsep=2pt,topsep=0pt,parsep=0pt]
    \item Context $\boldsymbol{S}=\{s_t\}_{t=1}^{N_t}$ in textual modality, which represents a series of coherent textual instructions to perform a specific skill or task (e.g., multiple steps to complete a recipe or a craft product).
    \item Question $\boldsymbol{Q}$ and Answer $\boldsymbol{A}$, which is either a single image or a series of images %
    in a reasoning task (e.g., visual cloze, visual coherence, or visual ordering). 
\end{itemize}
Following~\cite{DBLP:conf/sigir/LiuYZLLBX20}, we combine the images contained in %
$\boldsymbol{Q}$ and %
$\boldsymbol{A}$ to form $N_c$ candidate image sequences $\{a_1,...a_j,...a_{N_c}\}$. 
Let $N_a$ be the length of the $j$-th candidate image sequence $a_j=\{I_{j,1},...I_{j,N_a}\}$.
Take the visual cloze task as an example, we fill the placeholder of the question with candidate answers to form $N_c$ image sequences with length $N_a = 4$.
The model requires to select the most relevant candidate by calculating the similarity between text sequence $\boldsymbol{S}=\{s_t\}_{t=1}^{N_t}$ and each image sequence $a_j$.

\subsection{Homogeneous Graph Construction}\label{sec:graph construction}

As shown in Figure~\ref{fig:framework}, we first extract the tokens (objects) in text (image) as the initial nodes of homogeneous graph, respectively.

\noindent \textbf{Textual Node.}
Let $N_t$ be the number of textual instructions $\boldsymbol{S}=\{s_t\}_{t=1}^{N_t}$. 
First, each instruction $s_t$ is tokenized into a token sequence  $\{e_{t_{\text{[CLS]}}}, e_{t_1},...e_{t_\text{[SEP]}}\}$, where [CLS] and [SEP] are the special tokens introduced to mark the start and the end of each instruction.
Then, we utilize an off-the-shelf POS tagger~\cite{DBLP:conf/coling/AkbikBV18} to identify all nouns and noun phrases in the token sequence.
Finally, we concatenate all the token sequences of textual instructions $\boldsymbol{S}$ and feed the token embedding into the graph aggregation module.

\noindent \textbf{Visual Node.}
For each image sequence $a_j$, we employ a pre-trained Faster-RCNN to extract a set $\{e_{v_{\text{[CLS]}}}, e_{v_1},...e_{v_k}\}$ with $k$ object features as visual tokens.
Following~\cite{DBLP:conf/icpr/MessinaFEA20, DBLP:conf/iclr/DosovitskiyB0WZ21}, we reserve [CLS] as the beginning token for each image whose final embedding is regarded as the representation of the whole image.
The operation of visual node is in a similar manner as textual and any two nodes in the same instruction~(image) are connected to construct a homogeneous graph.

\subsection{Heterogeneous Graph Construction}\label{sec:edge construction}
Based on the homogeneous graph of each textual instruction~(image), we introduce various types of edges to construct our heterogeneous graph \name{}.

\subsubsection{Temporal Edge} \label{sec:3.3.1}
It is essential to model the temporal evolution of entities for comprehending procedural content.
Let us revisit the example in Figure~\ref{fig:recipe}. When a human reads step 4, the connection between entities (e.g., \textsf{strawberries} and \textsf{oranges}) and their descriptions in step 1 is naturally established.  %

We design the temporal edge to model the evolution of entities in text and image.
It can be seen that the temporal edge describes the evolution %
at different steps. For the textual nodes, the same entity %
appearing in different steps are connected by a textual temporal edge~(node-based).
While for the visual nodes, we directly calculate the Euclidean Distance between object features due to the absence of accurate object detection.
Following~\cite{song2021spatial}, if the distance between node (object) $i$ and node (object) $j$ is less than a threshold $\lambda_t$, we treat them as the same object and connect node $i$ to node $j$ via a visual temporal edge~(node-based).
Meanwhile, we consider that there may also be temporal evolution for edges, such as changes in the relationship between entities. Therefore, we also introduce temporal edge (edge-based) to characterize the change of edges.

\subsubsection{Modal Edge}\label{sec:3.3.2}
As shown in Figure~\ref{fig:recipe}, the textual instruction of each image can be viewed as a noisy form of image annotation~\cite{DBLP:conf/emnlp/HesselLM19}.
The association between image and sentence can be inferred through entity representations under different modalities.
Correspondingly, we design the intra-modal edge and the inter-modal edge to represent the modal interactions.
In Section~\ref{sec:graph construction}, any two nodes in the same modality and the same instruction (image) are connected by an intra-modal edge.
It is worth noting that for each instruction (image), the special [CLS] node is connected to all other nodes in order to aggregate graph-level features.

On the other hand, the textual node representing any entity and the corresponding visual node are connected by an inter-modal edge. 
We employ a visual grounding toolkit~\cite{DBLP:conf/iccv/YangGWHYL19} to detect visual objects for each noun phrase.
Specifically, we predict the bounding box corresponding to the text entity and compute the Intersection over Union (IoU) between all visual objects.
If the IoU between the prediction box and the visual box exceeds a threshold $\lambda_m$, the textual node and the corresponding visual node are connected by an inter-modal edge~(node-based).
Similar to section~\ref{sec:3.3.1}, considering the influence of entity-relationship under different modalities, we also introduce inter-modal edge~(edge-based) to characterize the interaction between edges.

\subsection{Graph Aggregation}\label{sec:graph aggregation}

\subsubsection{Node Encoding}
As described in Section 3.2, we have obtained the embeddings of the textual tokens and visual objects.
Similar to~\cite{DBLP:conf/acl/LiYYHC20}, all embeddings are mapped to a set of initial node embeddings, and each node embedding is the sum of 1) a textual token embedding or visual object embedding; 2) a position embedding that identifies the position of the token or object in the image\footnote{We exploit the bounding box feature extracted by Faster-RCNN as the position embedding of the object.}; and 3) a segment embedding generated from the step number in PMD which indicates different textual instructions or images.

\subsubsection{Edge Encoding}
To encode the structural information into \name{}, we consider the temporal encoding and the modal encoding separately. 
For any two nodes $v_i$ and $v_j$ in \name{}, we construct two mappings: $\phi_{t}(v_i,v_j)\rightarrow\mathbb R$ and $\phi_{m}(v_i,v_j)\rightarrow\mathbb R$  which encode the temporal edge and the modal edge between them.
The temporal encoding and the modal encoding of the total graph are fed into the graph-based aggregation module.

\subsubsection{Graph-Based Fusion}

As shown in the right part of Figure 2, we first introduce two multi-layer perceptrons (MLP) with Tanh activation function to project different node embeddings from two modalities into the same space.
Then, we extend the VisualBERT~\cite{DBLP:conf/acl/LiYYHC20} to the graph-based fusion layer, which %
concatenates the node embeddings from MLPs as input and outputs their graph-based joint representations.
Specifically, in each fusion layer, %
updating the hidden states of textual node and visual node mainly involve the following steps.

Firstly, we exploit a graph-based multi-head attention mechanism to generate contextual representations of nodes.
Formally, the output of the $h$-th attention head in the $l-1$ layer can be obtained as follows:
\begin{equation}
\setlength{\abovedisplayskip}{5pt} 
\setlength{\belowdisplayskip}{5pt}
    A\left(\boldsymbol q,\boldsymbol k,\boldsymbol v\right)^{h, l-1}_j=\sum_{i=1}^{N}\boldsymbol v_i^h\left(\text{\textit{Softmax}}(\boldsymbol e_{i,j}^h)\right),
    \label{eq:attention output}
\end{equation}
\begin{equation}
\setlength{\abovedisplayskip}{5pt} 
\setlength{\belowdisplayskip}{5pt}
    \boldsymbol e_{i,j}^h=\frac{\boldsymbol q_j^h\left(\boldsymbol k_i^h\right)^T}{\sqrt{\boldsymbol d}}+\boldsymbol b^h_{\phi_{m}\left(i,j\right)}+\boldsymbol b^h_{\phi_{t}\left(i,j\right)},
    \label{eq:attention score}
\end{equation}
where $q,k,v$ are the query matrix, key matrix, and value matrix generated from the hidden state $H^{(l-1)}$ of nodes in the $l-1$ layer. $\phi_{t}(i,j)$ and $\phi_{m}(i,j)$ denote the temporal encoding and the modal encoding of \name{}, which serve as bias terms in the attention module.
It is worth noting that each head in the multi-head attention mechanism exhibits a broad range of behaviors~\cite{DBLP:conf/acl/Vig19}; thus, we add different temporal encoding and modal encoding separately for each attention head.
Meanwhile, in order to model the relationship of edges, the temporal encoding and the modal encoding are learned separately for each layer.

\begin{algorithm}[t]

\small
\SetKwFunction{AddNorm}{Add\&Norm}
\SetKwInOut{Input}{Input}
\SetKwInOut{Output}{Output}
\normalem
\Input{The initial hidden states of \name{} $\boldsymbol H^0;$ the graph-based fusion layer number $\boldsymbol N$; the temporal encoding $\boldsymbol \Phi_t;$ the modal encoding $\boldsymbol \Phi_m;$ the attention head $\boldsymbol A$}
\Output{The final hidden states $\boldsymbol H^N$}
\For{$\boldsymbol l\leftarrow 1$ \KwTo $\boldsymbol N$}{
        \tcp{Graph-Based Fusion Layer}
        \ForEach{attention head $\boldsymbol A^{h,l}$ in layer $\boldsymbol l$}{
            Generate $\boldsymbol q,\boldsymbol k,\boldsymbol v$ from hidden states $\boldsymbol H^{l-1};$ \\
            Obtain edge encoding $\boldsymbol b^h_{\phi_{t}(i,j)}$ and $\boldsymbol b^h_{\phi_{m}(i,j)};$ \\
            Calculate attention weight $\boldsymbol e_{i,j}^h$ (Eq.~\ref{eq:attention score}) with $\{\boldsymbol q,\boldsymbol k, \boldsymbol b^h_{\phi_{t}(i,j)},\boldsymbol b^h_{\phi_{m}(i,j)}\};$\\
            Acquire output of head $\boldsymbol A^{h,l}$ (Eq.~\ref{eq:attention output}) with $\{\boldsymbol v,\boldsymbol e_{*,j}^h\};$\\
        }
    Aggregate each attention head $\boldsymbol A^h$\\
    \textbf{update} \\
    $\hat{\boldsymbol H}^{(l)}\leftarrow\text{LN}(W[\boldsymbol A^{1},...\boldsymbol A^{h}]+\boldsymbol H^{(l-1)});$ \\
    $\boldsymbol H^{(l)}\leftarrow\text{LN}(\text{FFN}(\hat{\boldsymbol H}^{(l)})+\hat{\boldsymbol H}^{(l)});$ (Eq.~\ref{eq:layer norm}) \\
}
return $\boldsymbol H^N$
\caption{\small{Graph Aggregation of \name{}}}
\label{al:algorithm graph}
\end{algorithm}
We concatenate the output of each head and pass them to a position-wise Feed Forward Networks (FFN) which is preceded and succeeded by residual connections and normalization layer (LN),
\begin{equation}
\setlength{\abovedisplayskip}{5pt} 
\setlength{\belowdisplayskip}{5pt}
\begin{aligned}
    \hat{\boldsymbol H}^{(l)}&=\text{LN}\left(W[\boldsymbol A^{1},...\boldsymbol A^{h}]+\boldsymbol H^{(l-1)}\right), \\
    \boldsymbol H^{(l)}&=\text{LN}\left(\text{FFN}(\hat{\boldsymbol H}^{(l)})+\hat{\boldsymbol H}^{(l)}\right),
    \label{eq:layer norm}
\end{aligned}
\end{equation}

where $W$ is a learnable parameter and $[\ ]$ denotes the concatenation manipulation.
Finally, based on \name{}, we stack multi-layer graph-based fusion layers to conduct graph encoding. Algorithm~\ref{al:algorithm graph} shows the aggregation of \name{} in detail.

\subsubsection{Graph-Based Reasoning}\label{sec:graph-based reasoning}
As mentioned in Section ~\ref{sec:graph construction}, we regard the hidden state of [CLS] as the representations of each instruction (image), where their final hidden states $H^T$ and $H^V$ are passed into the graph reasoning module for task completion.

Firstly, we leverage the one-to-one correspondence between instruction and image, e.g., each instruction has an image to visualize it~\cite{DBLP:journals/corr/abs-2109-11047}.
\name{} involves a Contrastive Coherence Loss for keeping the alignment between instruction and image.
Let $H^{V^+}$ and $H^{V^-}$ represent the positive and negative examples, the loss $\mathcal{L}^{\text{Coh}}$ of the $i$-th step can be defined as follows:
\begin{equation}
\setlength{\abovedisplayskip}{5pt} 
\setlength{\belowdisplayskip}{5pt}
    \label{eq:contrastive loss}
    \mathcal{\boldsymbol L}_{i}^{\text{Coh}}=-\log\frac{\exp\{sim(\boldsymbol H^{T}_i, \boldsymbol H^{V^+}_i)/\boldsymbol \tau\}}{\sum_{j=1}^K\exp\{sim(\boldsymbol H^{T}_i, \boldsymbol H^{V^-}_j)/\boldsymbol \tau\}},
\end{equation}
where $K$ is the total number of negative samples~\cite{DBLP:conf/cvpr/He0WXG20} generated from the min-batch, $sim(\cdot ,\cdot)$ and $\tau$ are the standard cosine similarity function and temperature.

In a downstream reasoning task, the model needs to predict the correct candidate $a_j=\{I_{j,1},...I_{j,N_a}\}$ based on the instructions $\boldsymbol{S}=\{s_t\}_{t=1}^{N_t}$.
Referring to the sentence image prediction task in ~\cite{DBLP:conf/acl/LiYYHC20}, we concatenate all representations of each candidate image sequence to generate a instruction candidate pair as:
$(\boldsymbol{S},a_j)=[\text{CLS}, H^T_1,...H^T_{N_t}, \text{SEP}, H^V_{j,1},...H^V_{j,{N_a}}]$,
where [CLS] and [SEP] are special tokens as used in ~\cite{DBLP:conf/acl/LiYYHC20}.
We pass this input through a shallow transformer followed by a fully connected layer to obtain the prediction score $P(\boldsymbol{S},a_j)$ for the $j$-th candidate, and the prediction loss can be defined as
\begin{equation}
\setlength{\abovedisplayskip}{5pt} 
\setlength{\belowdisplayskip}{5pt}
    \label{eq:predict loss}
    \mathcal{\boldsymbol L}^{\text{Pre}}=-\log\frac{\exp\left(P\left(\boldsymbol S,\boldsymbol a_j\right)\right)}{\sum_{i=1,i\neq j}^{N_a-1}\exp\left(P\left(\boldsymbol S,\boldsymbol a_i\right)\right)},
\end{equation}
where $a_j$ is the correct candidate and $N_a$ is the number of candidates.
We %
get the final loss function and optimize it through the Adam optimizer:
\begin{equation}
\setlength{\abovedisplayskip}{5pt} 
\setlength{\belowdisplayskip}{5pt}
    \label{eq:final loss}
    \mathcal{\boldsymbol L}=\mathcal{\boldsymbol L}^{\text{Pre}}+\lambda_b\mathcal{\boldsymbol L}^{\text{Coh}},
\end{equation}
where $\lambda_b$ is the balance parameter. 
Unless otherwise specified, all the results in this paper use $\lambda_b=0.1$ which we find to perform best.

\section{Experiments}

\subsection{Datasets and Metrics}
\noindent \textbf{RecipeQA.} RecipeQA~\cite{DBLP:conf/emnlp/YagciogluEEI18} is a multimodal comprehension dataset with 20K recipes approximately and more than 36K question-answer pairs.
Unlike other multimodal reading comprehension datasets~\cite{DBLP:conf/cvpr/TapaswiZSTUF16, DBLP:conf/cvpr/IyyerMGVBDD17, DBLP:conf/cvpr/KembhaviSSCFH17} analyze against movie clips or comics, RecipeQA requires reasoning %
real-world cases.

\noindent \textbf{CraftQA.} %
We collect CraftQA from Instructables\footnote{https://www.instructables.com/}, which is an online community where people can share their tutorials %
for accomplishing a task in a step-by-step manner.
Specifically, we collect the most visited tutorials and remove those that contain only text or video.
For question and answer generation, we also remove the tutorials that contain less than 3 images.
To construct the distractor of each task, we compute the Euclidean distance between the image features that are extracted from a pretrained ResNet-50~\cite{DBLP:conf/cvpr/HeZRS16}.
Taking the visual cloze task as an example, the distractor is sampled from the nearest neighbors of the ground-truth image based on Euclidean distance.
Finally, CraftQA contains about 27k craft product-making tutorials and 46k question-answer pairs.
We employ CraftQA to evaluate the reading comprehension performance of \name{} in different domains as well as its domain transfer capability.
More statistics about these two datasets are shown in Table~\ref{tab:dataset_label}. 

\noindent \textbf{Metric.} In three Procedural M$^3$C tasks that are tested in the following experiments (visual cloze, visual coherence, and visual ordering),
we use classification accuracy as %
the evaluation metric, which is defined as the percentage of yielding the ground-truth answer during testing~\cite{DBLP:conf/emnlp/YagciogluEEI18, DBLP:conf/conll/AmacYEE19, DBLP:conf/sigir/LiuYZLLBX20}.

\begin{table}
\centering

\small
\resizebox{0.48\textwidth}{!}{
\begin{tabular}{c|l|lll}
\toprule
\textbf{Dataset} & \textbf{Statistics} & \textbf{Train} & \textbf{Valid} & \textbf{Test} \\
\cmidrule{1-5}
\multirow{4}{*}{RecipeQA} & \# of recipes & 15,847 & 1,963 & 1,969 \\
 & avg. \# of steps & 5.99 & 6.01 & 6.00 \\
 & avg. \# of words & 443.01 & 440.51 & 435.33 \\
 & avg. \# of images & 12.67 & 12.74 & 12.65 \\
\cmidrule{1-5}
\multirow{4}{*}{CraftQA} & \# of tutorials & 21,790 & 2,601 & 2,604 \\
& avg. \# of steps & 7.53 & 7.47 & 7.56 \\
& avg. \# of words & 535.88 & 531.01 & 541.97 \\
& avg. \# of images & 20.14 & 20.19 & 20.37 \\
\bottomrule
\end{tabular}}
\caption{Statistics of RecipeQA and CraftQA dataset. Each dataset is split into training, validation, and test sets based on the number of recipes or craft-making tutorials. We also provide the average count of steps, images, and words contained in each split dataset.
}
\label{tab:dataset_label}
\end{table}

\begin{table*}
\centering
\small
\resizebox{0.98\textwidth}{!}{
\begin{tabular}{p{4.0cm}| p{1.0cm}<{\centering} p{1.2cm}<{\centering} p{1.0cm}<{\centering} p{1.0cm}<{\centering}| p{1.0cm}<{\centering} p{1.2cm}<{\centering} p{1.0cm}<{\centering} p{1.0cm}<{\centering}}
\toprule
 & \multicolumn{4}{c|}{\textit{RecipeQA}} & \multicolumn{4}{c}{\textit{CraftQA}} \\
\cmidrule{2-9}
Model & Cloze & Coherence & Ordering & Average & Cloze & Coherence & Ordering & Average \\
\cmidrule{1-9}
\morecmidrules\cmidrule{1-9}
Human~\cite{DBLP:conf/conll/AmacYEE19}                      & 77.60 & 81.60 & 64.00 & 74.40 & 55.33 & 61.80 & 54.60 & 57.24 \\
\cmidrule{1-9}
HS~\cite{DBLP:conf/emnlp/YagciogluEEI18}               & 27.35 & 65.80 & 40.88 & 44.68 & 27.81 & 42.61 & 35.23 & 35.22 \\
PRN~\cite{DBLP:conf/conll/AmacYEE19}            & 56.31 & 53.64 & 62.77 & 57.57 & 35.14 & 33.67 & 42.31 & 37.04 \\
MLMM-Trans~\cite{DBLP:conf/sigir/LiuYZLLBX20}                  & 65.57 & 67.33 & 63.75 & 65.55 & 39.43 & 46.52 & 43.62 & 43.20 \\
VisualBERT~\cite{DBLP:conf/acl/LiYYHC20}                        & 66.49 & 68.15 & 63.09 & 65.91 & 40.56 & 47.84 & 46.18 & 44.86 \\
\cmidrule{1-9}
\rowcolor{gray!25}
\name{}~(Our Model)                 & \textbf{73.27} & \textbf{70.38} & \textbf{65.56} & \textbf{69.73} & \textbf{48.01} & \textbf{51.36} & \textbf{49.25} & \textbf{49.54} \\

\quad \textit{w/o} Temporal Encoding                 & 70.18 & 68.81 & 63.91 & 67.63 & 45.91 & 47.71 & 47.24 & 46.95 \\
\quad \textit{w/o} Modal Encoding                 & 71.50 & 69.63 & 64.15 & 68.42 & 46.67 & 49.06 & 48.35 & 48.02 \\
\quad \textit{w/o} Both Encoding                 & 68.97 & 67.72 & 63.34 & 66.67 & 44.87 & 46.29 & 46.77 & 45.98 \\
\quad \textit{w/o} Contrastive Loss $\mathcal{L}^{Coh}$                 & 71.66 & 68.54 & 64.54 & 68.24 & 47.49 & 50.03 & 48.79 & 48.77 \\

\bottomrule
\end{tabular}}
\caption{Experimental comparison of procedural multimodal machine comprehension on RecipeQA and CraftQA: 
``\textit{w/o} Temporal Encoding'' and ``\textit{w/o} Modal Encoding'' denote to remove the temporal or modal edges respectively and ``\textit{w/o} Both Encoding'' denotes to remove both of them. 
``\textit{w/o} Contrastive Loss $\mathcal{L}^{Coh}$'' denotes excluding the contrastive coherence loss $\mathcal{L}^{Coh}$ from the final loss. Similar to RecipeQA, 100 questions in CraftQA are extracted from its validation set to evaluate the ``Human'' performance.}
    \label{tab:main_label}
\end{table*}

\subsection{Implementation Details}
For visual node construction, we employ the pre-trained Faster R-CNN~\cite{DBLP:conf/nips/RenHGS15} model provided by Detectron2~\cite{wu2019detectron2} and limit the number of objects to 36 for each image.
Following~\cite{DBLP:conf/iccv/YangGWHYL19,song2021spatial}, we set the thresholds $\lambda_t$ and $\lambda_m$ as 7 and 0.5, respectively, for the temporal and the modal edge constructions.

The framework of the graph-based fusion module is built on VisualBERT~\cite{DBLP:conf/acl/LiYYHC20} with its initialized parameters and tokenizer implemented by HuggingFace's transformers library~\cite{DBLP:conf/emnlp/WolfDSCDMCRLFDS20}.
The shallow transformer in the graph-based reasoning module is designed as 2 hidden layers with a size of 512 and 8 attention heads.
During the training stage, the batch size is fixed to 16 and the number of negative samples $K$ is set to 8.
The temperature parameter $\tau$ in Eq.(\ref{eq:contrastive loss}) is set to 0.07.
The balance parameter $\lambda_b$ in Eq.(\ref{eq:final loss}) is set to 0.1.
Adam with the learning rate $5\times 10^{-5}$ is used to update parameters. 
We introduce an early stopping mechanism and set the patience value to 5, which means the training will stop if the model performance is not improved in five consecutive times.
Our source code will be %
released online. %

\subsection{Baselines}
We compare our model with the following models: 
(1) \textbf{Hasty Student (HS)}~\cite{DBLP:conf/emnlp/YagciogluEEI18} discards %
textual context and directly exploits the similarities and dissimilarities between answer images to rank candidates.
(2) \textbf{PRN}~\cite{DBLP:conf/conll/AmacYEE19} introduces external relational memory units to keep track of textual entities and employs a bi-directional attention mechanism to obtain a question-aware embedding for prediction. 
(3) \textbf{MLMM-Trans}~\cite{DBLP:conf/sigir/LiuYZLLBX20} modifies the framework of the transformer~\cite{DBLP:conf/nips/VaswaniSPUJGKP17} and conducts an intensive attention mechanism at multiple levels to predict correct image sequences.
(4) \textbf{VisualBERT}~\cite{DBLP:conf/acl/LiYYHC20} consists of a stack of transformer layers that extend the traditional BERT~\cite{DBLP:conf/naacl/DevlinCLT19} model to a multimodal encoder.
The performance of some baselines on RecipeQA has been previously reported in ~\cite{DBLP:conf/conll/AmacYEE19, DBLP:conf/sigir/LiuYZLLBX20}.

\subsection{Experimental Results}
\subsubsection{Comparison %
Analysis}
As shown in Table~\ref{tab:main_label},
\name{} shows favorable performance in different reasoning tasks, with an average accuracy of 69.73 and 49.54 on RecipeQA and CraftQA, following behind the ``Human'' performance.
Besides, the performance on the \textit{visual ordering} task exceeds human accuracy for the first time, which proves that the temporal and modal analysis in \name{} is effective in comprehending PMDs. 
MLMM-Trans performs comparably with VisualBERT while inferior to \name{}, which may be attributed to their superficial consideration of entities.

MLMM-Trans %
ignores the entity information contained in text (e.g., the correspondence between entities in text and images) and
VisualBERT directly fuses textual and visual features without considering %
entity evolution. 
In \name{}, we explicitly identify and model entity evolution in PMD, whereas MLMM-Trans and VisualBERT assume entity information to be learned implicitly alongside other data.

Meanwhile, CraftQA has more images (20.14\,vs\,12.67) and tokens (535.88\,vs\,443.01) on average than RecipeQA. More diverse complex cases in CraftQA require better comprehension and reasoning capacities for both models and humans. We believe this can explain the lower results on CraftQA.
This emphasizes the necessity of comprehending entity coherence in a multimodal context.
\begin{figure}
    \begin{minipage}{1\linewidth}
    \centering
    \includegraphics[width=\linewidth]{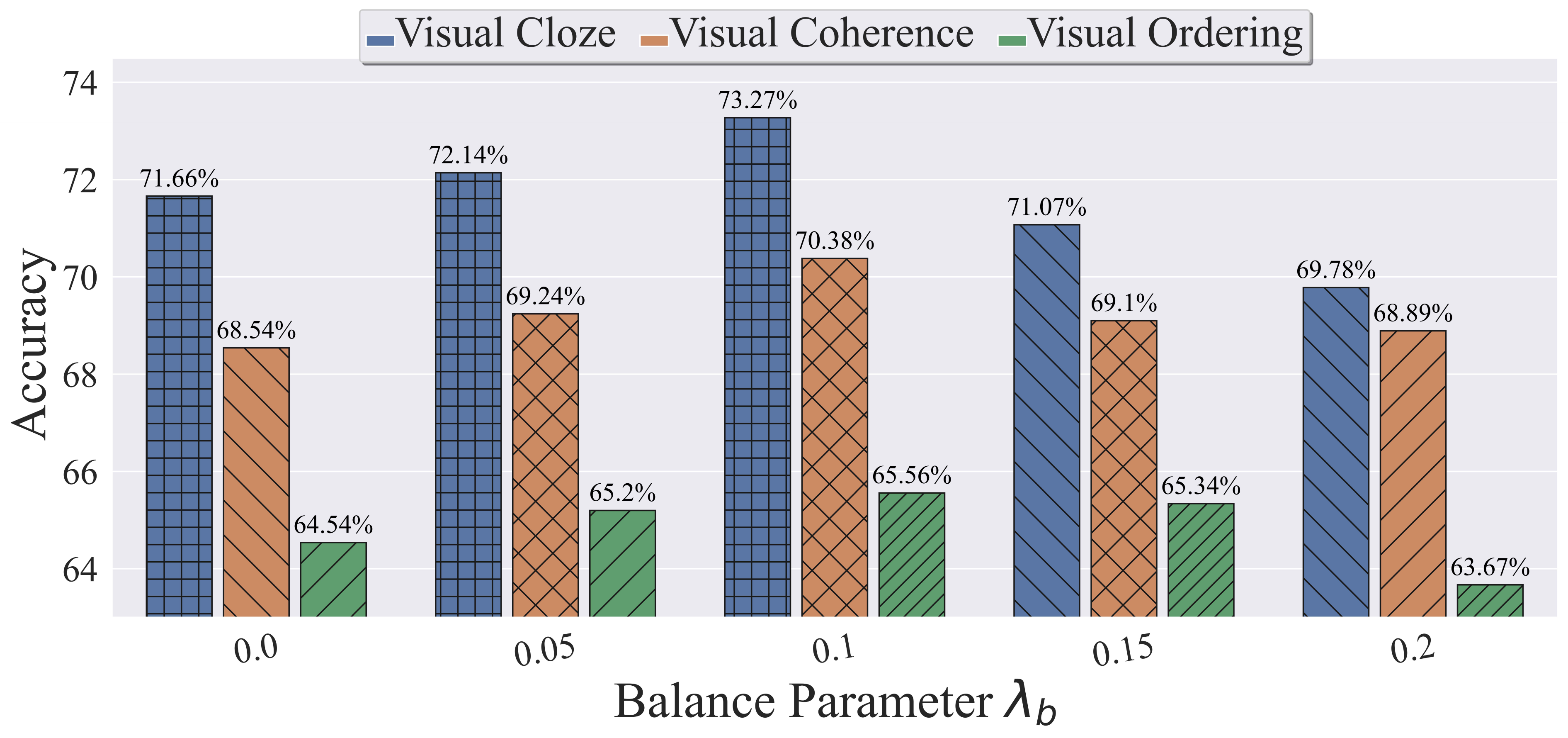}
    \end{minipage}
\caption{Experimental results of our model with different balance parameter $\lambda_b$ in Eq.(\ref{eq:final loss}) on RecipeQA.}
\label{fig:parameter_study}
\end{figure}

\begin{figure*}[htb]
  \centering
  \includegraphics[width=1\textwidth]{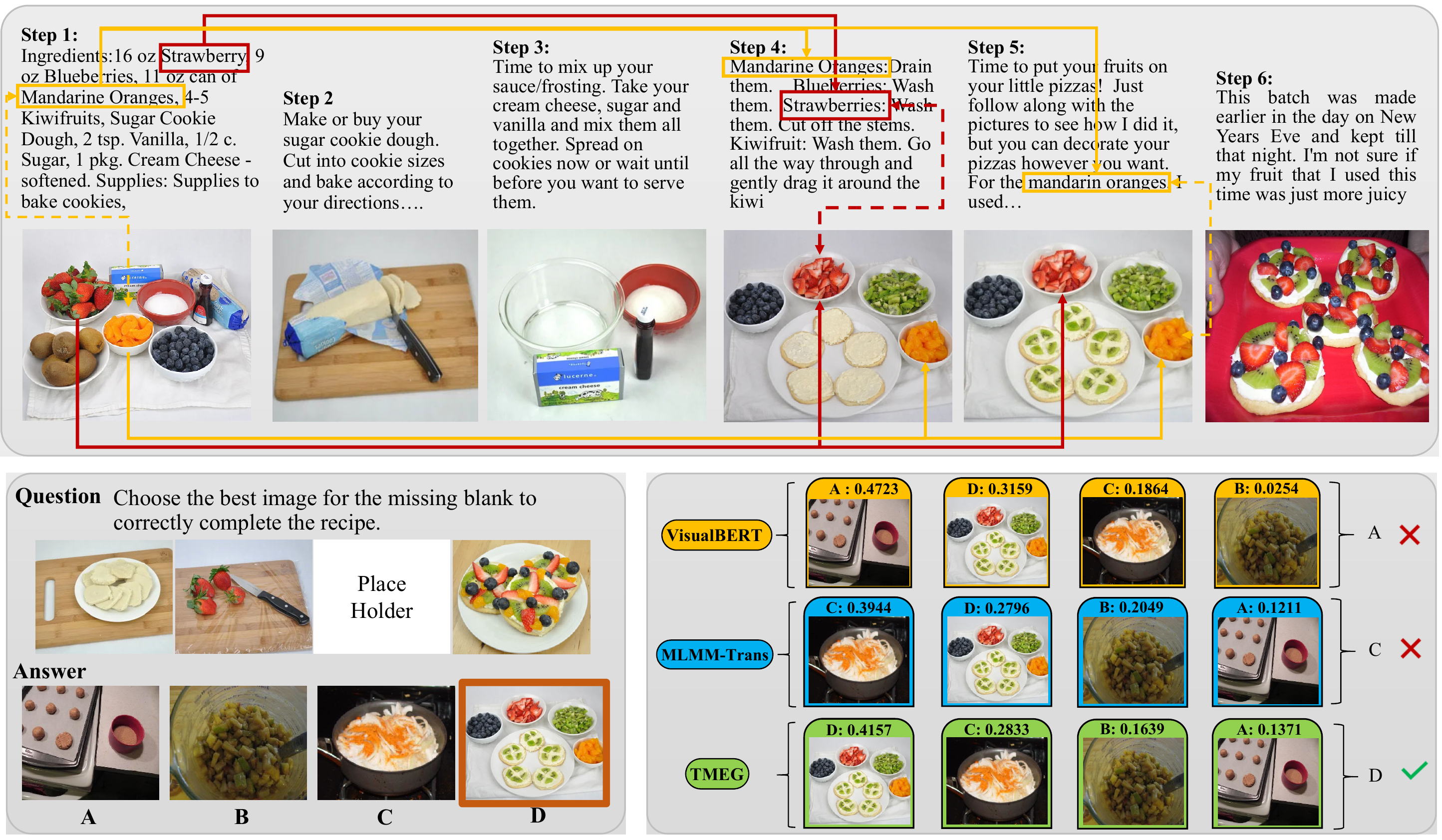}\\
                    \caption{An illustrated example of the visual cloze task on RecipeQA to clarify the workflow of \name{} when dealing with the downstream tasks.}
  \label{fig:case_study}
\end{figure*}

\subsubsection{Ablation Study}
We evaluate the effects of temporal encoding, %
modal encoding, %
and contrastive coherence loss $\mathcal{L}^{\text{Coh}}$ to examine the three modules.

\noindent
\textbf{Edge Encoding.}
Table~\ref{tab:main_label} also shows the ablation results of our model when each module is respectively removed. %
In terms of edge encoding, removing temporal encoding %
makes more negative effects on \name{} than moving the modal encoding, %
reflecting the significance of modeling temporal entity evolution for procedural M$^3$C.

\noindent
\textbf{Contrastive Coherence Loss.} 
As shown in the last row of Table~\ref{tab:main_label}, we find that $\mathcal{L}^{\text{Coh}}$ can indeed improve %
\name{}. 
The reason is that $\mathcal{L}^{\text{Coh}}$ helps enhance the learning of textual entity representation and visual entity representation in Procedural M$^3$C.

\subsection{Analysis of \name{}}

\subsubsection{Balance Parameter} 
In Figure~\ref{fig:parameter_study}, we illustrate the influence of the balance parameter $\lambda_b$ in Eq.(\ref{eq:final loss}), which balances the contrastive coherence loss $\mathcal{L}^{\text{Coh}}$ and the candidate prediction loss $\mathcal{L}^{\text{Pre}}$.
We tune $\lambda_b$ from 0 to 0.2 with 0.05 as the step size.
We observe that %
the model beats the highest accuracy when $\lambda_b=0.1$.
Generally,
(1) introducing the contrastive coherence loss %
can improve \name{} for better fitting downstream tasks, and
(2) appropriately balancing %
the prediction loss $\mathcal{L}^{\text{Pre}}$ and contrastive coherence loss $\mathcal{L}^{\text{Coh}}$ helps \name{} comprehend PMDs.

\begin{table}
\small
\centering
\resizebox{0.48\textwidth}{!}{
    \begin{tabular}{
        p{4.0cm}<{}|
        p{0.8cm}<{\centering} 
        p{0.8cm}<{\centering}
        p{0.8cm}<{\centering}}
        \toprule
        Model & R2C & C2R & Average \\
        \cmidrule{1-4}
        MLMM-Trans~\cite{DBLP:conf/sigir/LiuYZLLBX20} & 33.98 & 40.14 & 37.06 \\
        VisualBERT~\cite{DBLP:conf/acl/LiYYHC20}    & 35.24 & 42.15 & 38.69 \\
        \rowcolor{gray!25}
        \name{}~(Our Model)     & \textbf{39.06} & \textbf{45.42} & \textbf{42.24} \\
        \name{}~(\textit{w/o} Edge Encoding)     & 36.02 & 43.17 & 39.60 \\
    \bottomrule
    \end{tabular}
}
    \caption{Results of the domain transfer experiments, where ``R2C'' denotes training models on RecipeQA while testing on CraftQA, and ``C2R'' represents the opposite.}
    \label{tab:domain transfer}
\end{table}

\subsubsection{Cross-Domain Investigation}
To study the domain transfer capability of our framework, we %
evaluate \name{} in different domains, as shown in Table~\ref{tab:domain transfer}.
Specifically, The model trained on RecipeQA is evaluated on CraftQA, and the reverse is true for CraftQA.
Results show that compared with other baselines, our model achieves more generalized and better comprehension performance on domain transfer by incorporating \name{}.
\subsubsection{Case Study}
Figure~\ref{fig:case_study} further presents a visual cloze example on RecipeQA which requires a correct image in the missing piece after reading the context.
We compare the highest-scored candidate images respectively picked out by MLMM-Trans~\cite{DBLP:conf/sigir/LiuYZLLBX20}, VisualBERT~\cite{DBLP:conf/acl/LiYYHC20}, and \name{}.
By considering the temporal-modal entity evolution, \name{} can capture the salient entities (e.g., \textsf{Strawberry} and \textsf{Sugar Cookie Dough}) and trace their evolution at each step, thereby inferring the ground-truth answer.

\section{Conclusion}
In this paper, we propose a novel temporal-modal entity graph (\name{}) to approach Procedural M$^3$C. 
Based on \name{}, we introduce graph-based fusion module and reasoning module, which are used to aggregate node features and solve downstream reasoning tasks.

What's more, we introduce another Procedural M$^3$C dataset called CraftQA to assist in evaluating the generalization performance of \name{} in different domains and domain transfer.
Extensive experiments on the RecipeQA and CraftQA validate the superiority of \name{}.
A promising future direction is to introduce temporal-modal entity graphs into the video understanding task~\cite{DBLP:conf/acl/LinRCNBDD20,DBLP:journals/corr/abs-2005-00706}, which also calls for an enhancement of the temporal and the cross-modal reasoning capability.
\section*{Ethical Considerations}

\noindent\textbf{Intellectual Property.}
CraftQA contains question answer pairs generated from copyright free tutorials found online\footnote{https://www.instructables.com}.
All of the tutorials are licensed with the Creative Commons license\footnote{https://creativecommons.org/licenses/by-nc-sa/4.0} which helps share knowledge and creativity for common use.

The collection of CraftQA is in accordance with the Terms of Service of Instructables as follows: by posting, providing, uploading, submitting, sharing, publishing, distributing, making available or allowing others to access and/or use Your Content to or through the Service You are solely responsible and liable for the consequences of doing so and you acknowledge and agree that Your Content can and may be viewed worldwide\footnote{https://www.autodesk.com/company/legal-notices-trademarks/terms-of-service-autodesk360-web-services/instructables-terms-of-service-june-5-2013}. %

We also construct experimental evaluations on the RecipeQA dataset. 
Referring to the official dataset descriptions of RecipeQA\footnote{https://hucvl.github.io/recipeqa/recipeqa-datasheet.pdf}, Legal and Ethical Considerations had been taken into account during the construction of RecipeQA. We have cited the corresponding papers in this study.

\noindent\textbf{Privacy.}
According to the Privacy
Statement of Instructables\footnote{https://www.autodesk.com/company/legal-notices-trademarks/privacy-statement}, users can choose whether or not to expose their information when publishing tutorials. Respecting personal privacy, we have removed all of the personal information of users from CraftQA and promise CraftQA isn't involved with any privacy issues.\\

\noindent \textbf{Acknowledgements}: This work was supported in part by the National Natural Science Foundation of China (No. 62106091) and Shandong Provincial Natural Science Foundation (No. ZR2021MF054).

\normalem
\bibliography{main.bbl}
\bibliographystyle{acl_natbib}

\end{document}